\documentclass[letterpaper]{article} % DO NOT CHANGE THIS
\usepackage{aaai2026}
\usepackage{times}  % DO NOT CHANGE THIS
\usepackage{helvet}  % DO NOT CHANGE THIS
\usepackage{courier}  % DO NOT CHANGE THIS
\usepackage[hyphens]{url}  % DO NOT CHANGE THIS
\usepackage{graphicx} % DO NOT CHANGE THIS
\usepackage{natbib}  % DO NOT CHANGE THIS AND DO NOT ADD ANY OPTIONS TO IT
\usepackage{caption} % DO NOT CHANGE THIS AND DO NOT ADD ANY OPTIONS TO IT
\usepackage{algorithm}
\usepackage{algorithmic}
\usepackage{amsmath}
\usepackage{makecell}
\usepackage{multirow}
\usepackage{booktabs}
\usepackage{tabularx}
\usepackage{enumitem}
\usepackage{colortbl}       % 专门用于表格颜色控制
\usepackage{arydshln}
\usepackage{amssymb}
\usepackage{newfloat}
\usepackage{listings}
\DeclareCaptionStyle{ruled}{labelfont=normalfont,labelsep=colon,strut=off} % DO NOT CHANGE THIS
\floatstyle{ruled}
\newfloat{listing}{tb}{lst}{}
\floatname{listing}{Listing}

\title{SlimInfer: Accelerating Long-Context LLM Inference via \\ Dynamic Token Pruning}
\author {
    Lingkun Long\textsuperscript{\rm 1},
    Rubing Yang\textsuperscript{\rm 1},
    Yushi Huang\textsuperscript{\rm 2},
    Desheng Hui\textsuperscript{\rm 1},
    Ao Zhou\textsuperscript{\rm 1}\footnotemark[1],
    Jianlei Yang\textsuperscript{\rm 1}\thanks{ Corresponding authors are Ao Zhou and Jianlei Yang. \\ \indent \indent Email: aozhou@buaa.edu.cn, jianlei@buaa.edu.cn. } % This work is supported in part by the National Natural Science Foundation of China (Grant No. 62572036), the Beijing Natural Science Foundation (Grant No. L243031), the National Key R\&D Program of China (Grant No. 2023YFB4503704 and 2024YFB4505601).}
}
\affiliations {
    \textsuperscript{\rm 1}Beihang University\\
    \textsuperscript{\rm 2}Hong Kong University of Science and Technology\\
}

\usepackage{bibentry}
\begin{document}

\newif\ifappendixonly
\appendixonlyfalse

\ifappendixonly
\else

\maketitle

\begin{abstract}
Long-context inference for Large Language Models (LLMs) is heavily limited by high computational demands.
While several existing methods optimize attention computation, they still process the full set of hidden states at each layer, limiting overall efficiency.
In this work, we propose SlimInfer, an innovative framework that aims to accelerate inference by directly pruning less critical prompt tokens during the forward pass.
Our key insight is an \textit{information diffusion phenomenon}: As information from critical tokens propagates through layers, it becomes distributed across the entire sequence. This diffusion process suggests that LLMs can maintain their semantic integrity when excessive tokens, even including these critical ones, are pruned in hidden states.
Motivated by this, SlimInfer introduces a dynamic fine-grained pruning mechanism that accurately removes redundant tokens of hidden state at intermediate layers. This layer-wise pruning naturally enables an asynchronous KV cache manager that prefetches required token blocks without complex predictors, reducing both memory usage and I/O costs.
Extensive experiments show that SlimInfer can achieve up to $\mathbf{2.53\times}$ time-to-first-token (TTFT) speedup and $\mathbf{1.88\times}$ end-to-end latency reduction for LLaMA3.1-8B-Instruct on a single RTX 4090, without sacrificing performance on LongBench. Our code will be available at https://github.com/Longxmas/SlimInfer.
\end{abstract}

\section{Introduction}\label{sec:intro}
Large Language Models (LLMs) have shown strong performance in long-context tasks such as summarization~\cite{zhang2020pegas, 2022booksum}, multi-document question answering~\cite{yang2018hotpotqa}, and retrieval from extended inputs~\cite{bai2024longbenchbilingualmultitaskbenchmark}. Scaling to longer sequences not only enables more complex reasoning, but also introduces substantial computational and memory overhead as the context length increases~\cite{fu2024challenges}.

\begin{figure}[t]
\centering
\includegraphics[width=1.0\columnwidth]{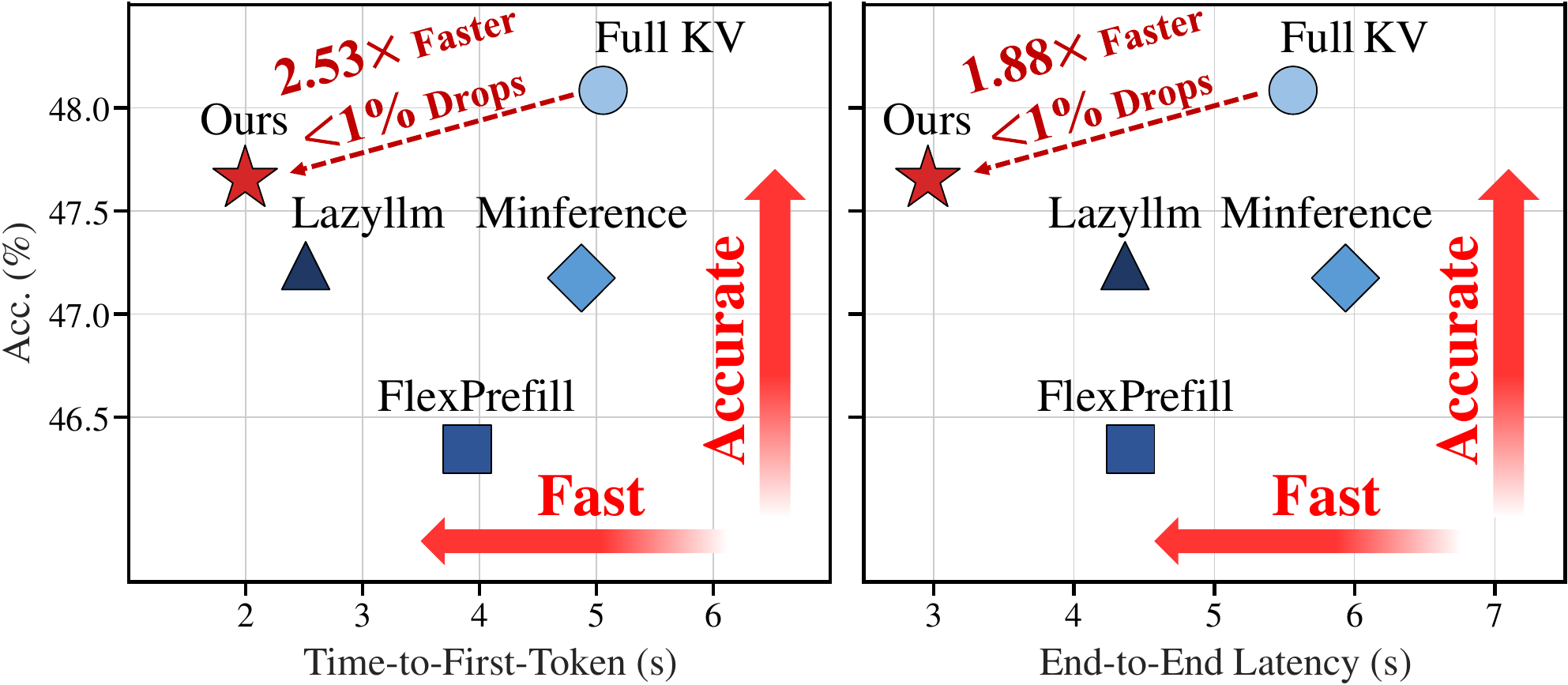}
\caption{Accuracy \emph{vs.} inference efficiency across different acceleration approaches on LongBench~\cite{bai2024longbenchbilingualmultitaskbenchmark} for LLaMA-3.1-8B-Instruct~\cite{grattafiori2024llama3herdmodels}.}
\label{fig:intro}

\end{figure}

During the prefill stage, the self-attention mechanism~\cite{attention2017} incurs quadratic time complexity with respect to the sequence length, making it a major source of latency in long-context scenarios. At the same time, the Key-Value (KV) cache grows linearly with input length, leading to substantial GPU memory consumption. To mitigate these issues, numerous token pruning methods have been proposed. However, existing token pruning methods face several critical limitations. Some works~\cite{zhang2023h2oheavyhitteroracleefficient, xiao2024efficientstreaminglanguagemodels, li2024snapkvllmknowslooking, yang2024tidaldecodefastaccuratellm, wang2025llmsknowdropselfattention, cai2025pyramidkvdynamickvcache, hao2025omnikv, nguyen2025fastgenfastcosteffectivesynthetic} focus primarily on optimizing the decoding phase, offering minimal improvements to the critical Time-To-First-Token (TTFT). In addition, their token eviction strategies often lead to accuracy degradation due to the removal of contextually important information. Other approaches~\cite{lai2025flexprefillcontextawaresparseattention, jiang2024minference10acceleratingprefilling} extend support to both prefill and decoding phases by sparsifying the attention pattern~\cite{deng2025sparseattention}. Nevertheless, they process the full sequence of hidden states at every layer, leaving non-attention components like Feed-Forward Networks (FFNs) unoptimized and limiting overall acceleration. Memory efficiency presents an additional challenge. Dynamic token pruning methods~\cite{fu2024lazyllmdynamictokenpruning} retain the entire KV cache in the GPU, leading to excessive memory consumption and limited scalability for longer sequences. To alleviate this, some systems offload the KV cache to the CPU~\cite{tang2024quest}, which reduces GPU pressure~\cite{gong2024llmcbenchmarkinglargelanguage, huang2024tfmqdmtemporalfeaturemaintenance} but introduces significant I/O latency. More recent designs attempt to prefetch KV segments to overlap data transfer and computation~\cite{lee2024infinigen, yang2025attentionpredictor}. However, these approaches often rely on predictor-based mechanisms, introducing additional overhead and complexity. Therefore, existing token pruning methods still struggle to simultaneously optimize inference speed~\cite{jiang2024minference10acceleratingprefilling, huang2025harmonicaharmonizingtraininginference, chen2025pre, chen2025tokenflowresponsivellmtext}, memory usage~\cite{xiao2024efficientstreaminglanguagemodels, huang2025qvgenpushinglimitquantized}, and model performance~\cite{huang2025temporalfeaturemattersframework, wnag2024ptsbenchcomprehensiveposttrainingsparsity}.

In this paper, we propose \textbf{SlimInfer}, a framework designed to accelerate inference by dynamically pruning less critical prompt tokens during the forward pass. Our method builds on a key insight we term the \textit{information diffusion phenomenon}: As information from critical tokens propagates through the layers of an LLM, it becomes progressively distributed across other token representations. This diffusion process suggests that LLMs can maintain their semantic integrity even when excessive tokens are pruned in hidden states, including those essential initially.  Motivated by this insight, SlimInfer introduces a dynamic layer-wise pruning to the hidden states across intermediate layers, progressively reducing computational workload. To preserve essential semantic information while maximising efficiency, we further introduce a fine-grained, block-wise importance evaluation that retains only the contextually relevant tokens. This pruning mechanism works in tandem with an asynchronous KV cache manager, which exploits the determinism of pruning decisions to enable predictor-free prefetching and efficient GPU memory management.

% % ZA：这一段和上面已经有部分重叠了，在这里单起一段介绍会打断整体叙事逻辑，建议提取凝练下关键信息合并到上一段落
% Motivated by these insights, SlimInfer introduces a dynamic pruning algorithm that operates at intermediate layers to progressively reduce the computational workload. To preserve essential information, this algorithm employs a fine-grained, block-wise importance evaluation strategy, ensuring that contextually critical tokens are retained. This pruner works in concert with an asynchronous KV cache manager that leverages the predictor-free prefetching opportunities to efficiently manage GPU memory and hide I/O latency.

We conduct extensive experiments on LLaMA-3.1-8B-Instruct~\cite{grattafiori2024llama3herdmodels} and Qwen2.5-7B-Instruct~\cite{qwen2025qwen25technicalreport}. As shown in Figure~\ref{fig:intro}, SlimInfer can achieve up to $2.53\times$ time-to-first-token (TTFT) speedup and a $1.88\times$ end-to-end latency reduction on a single NVIDIA RTX 4090 GPU. It also maintains near-lossless accuracy drops on the LongBench~\cite{bai2024longbenchbilingualmultitaskbenchmark}. 

\section{Related Works}
\subsection{Token Pruning}
Token pruning methods aim to reduce inference overhead by selectively removing less critical tokens from computation or memory. Many approaches target GPU memory reduction by maintaining a fixed-size KV cache. StreamingLLM~\cite{xiao2024efficientstreaminglanguagemodels} retains initial tokens (attention sinks) and a sliding window of recent tokens, but discards intermediate ones. H2O~\cite{zhang2023h2oheavyhitteroracleefficient} proposes a heavy-hitter oracle that evicts tokens with low cumulative attention scores. Similarly, SnapKV~\cite{li2024snapkvllmknowslooking} uses the local context of a prompt to predict and retain important tokens for future generation steps. LazyLLM~\cite{fu2024lazyllmdynamictokenpruning} introduces dynamic pruning based on token importance, but still retains most KV entries in GPU memory, limiting its scalability to longer contexts. A primary limitation of these methods is their irreversible token eviction, which permanently removes KV entries from GPU memory. This permanent removal can lead to significant accuracy degradation, particularly in complex tasks that rely on long-range dependencies scattered throughout the context. Unlike prior methods that irreversibly discard evicted tokens, SlimInfer offloads currently irrelevant tokens (\emph{i.e.}, pruned tokens) to CPU memory instead of discarding them, significantly improving performance and reducing GPU memory usage.
Other methods focus on accelerating computation by inducing sparsity in the attention map. FlexPrefill~\cite{lai2025flexprefillcontextawaresparseattention}, SpargeAttn~\cite{zhang2025spargeattention} adopt block-level heuristics by constructing representative vectors for token chunks, enabling coarse-grained attention skipping. In contrast, MInference~\cite{jiang2024minference10acceleratingprefilling} predicts structured sparse patterns based on partial attention observations. However, they still compute over the full sequence of hidden states at every layer. As a result, non-attention components like the Feed-Forward Networks (FFN) remain unoptimized, leaving significant room for further acceleration, which limits the overall speedup, especially during the prefill stage. SlimInfer directly addresses this by pruning the hidden states themselves, reducing the workload for all subsequent layers.

\subsection{KV Cache Offloading}

\begin{figure*}[t]
\centering
\includegraphics[width=0.9\textwidth]{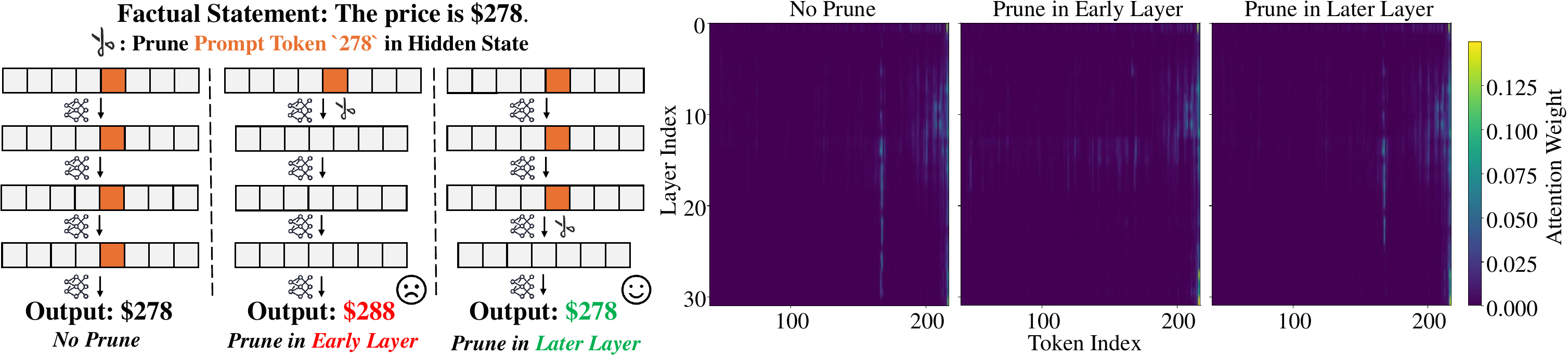} 
\caption{
(\textbf{Left}) Illustration of a probing experiment on LLaMA3.1-8B-Instruct~\cite{grattafiori2024llama3herdmodels}. Pruning the hidden state of the critical prompt token ``278'' (which indicates the correct answer: ``\$278'') in a later layer (right) results in the correct output, whereas pruning prompt tokens in an early layer (middle) leads to an incorrect output. 
(\textbf{Right}) Visualization of layer-wise attention weights from the decoding token (\emph{i.e.}, response token) ``278'' to prompt tokens.
}
\label{fig:motivation}

\end{figure*}

This line of work addresses the memory overhead of long-context inference by offloading the KV cache from GPU to CPU memory. Quest~\cite{tang2024quest} adopts a naive on-demand strategy, which fetches KV entries only when needed. More advanced systems attempt to prefetch KV cache blocks to overlap data transfer with computation. InfiniGen~\cite{lee2024infinigen} performs a lightweight rehearsal using partial model weights and previous-layer inputs, aided by offline Singular Value Decomposition (SVD). AttentionPredictor~\cite{yang2025attentionpredictor} trains a separate CNN to forecast attention scores. However, these approaches introduce considerable computational and engineering overhead. In contrast, SlimInfer sidesteps these limitations by leveraging its layer-wise pruning design to enable a predictor-free prefetching strategy, allowing efficient KV cache transfers without speculative estimation.
%  As shown in Figure~\ref{fig:kv_fetching}, this leads to I/O operations on the critical path and introduces substantial latency, particularly during the attention phase. In contrast, SlimInfer leverages deterministic pruning to prefetch critical KV blocks in advance, enabling better overlap between data transfer and computation.
\section{Motivation}\label{sec:motivation}

The design of SlimInfer is inspired by the following core insights: (1) \emph{Information diffusion phenomenon}, which confirms the feasibility of aggressive pruning hidden states; (2) This pruning strategy naturally offers an opportunity for KV cache prefetching to further improve inference efficiency.

% while the second reveals a unique system-level opportunity for I/O optimization that has been previously overlooked.

\subsection{Information Diffusion}\label{sec:inf}

Conventional token pruning approaches ~\cite{lai2025flexprefillcontextawaresparseattention, jiang2024minference10acceleratingprefilling} to accelerating attention computation typically retain the full set of hidden states while optimizing the underlying operations. In contrast, we investigate a more radical direction: The feasibility of pruning hidden states directly during the forward pass. To this end, we conducted a probing experiment on LLaMA3.1-8B-Instruct~\cite{grattafiori2024llama3herdmodels}. As shown in Figure~\ref{fig:motivation} (left), we selectively remove the hidden state corresponding to a critical prompt token ``278'' in different layers. The model successfully recalls the correct answer when pruning is applied at a later layer, but fails when pruning occurs earlier. To further understand the underlying mechanism, we visualize the attention weights from the decoding token to the prompt tokens across all transformer layers in Figure~\ref{fig:motivation} (right). In a standard decoding step, a bright vertical activation band emerges around Layer 13, which signifies a sustained focus of the decoding token towards the critical prompt token (``278'' in the response $\rightarrow$ ``278'' in the prompt). When pruning is applied in a later layer (\emph{i.e.}, Layer 25), the activation band is abruptly truncated at the pruning point. Despite this truncation, the model produces the correct output, suggesting that \textit{the semantic contribution of the critical token has already been effectively diffused into other tokens during the forward passes of the early layers.} To the contrary, early pruning at Layer 5 prevents the formation of this stable attention pattern. The absence of the hidden state corresponding to the critical token results in scattered and faint vertical lines over irrelevant tokens around Layer 13. This disoriented attention span reflects a disruption of the standard inference process.

% 对于hidden states来说，前面的层需要保留，剪后面的可以剪得多，原因是，重要的都可以剪，不重要的更加可以剪
\noindent\underline{\textit{Insights.}} This series of observations yields two core design principles for SlimInfer:
(\textit{i}) The hidden state of the early layer should be retained to preserve semantic fidelity, as pruning too early disrupts the diffusion process;
(\textit{ii}) In later layers, even the hidden states of originally important tokens can be safely pruned, indicating substantial redundancy that can be exploited to reduce computation.

\subsection{Prefetching Opportunities}\label{sec:pref}

\begin{figure}[t]
\centering
\includegraphics[width=1.0\columnwidth]{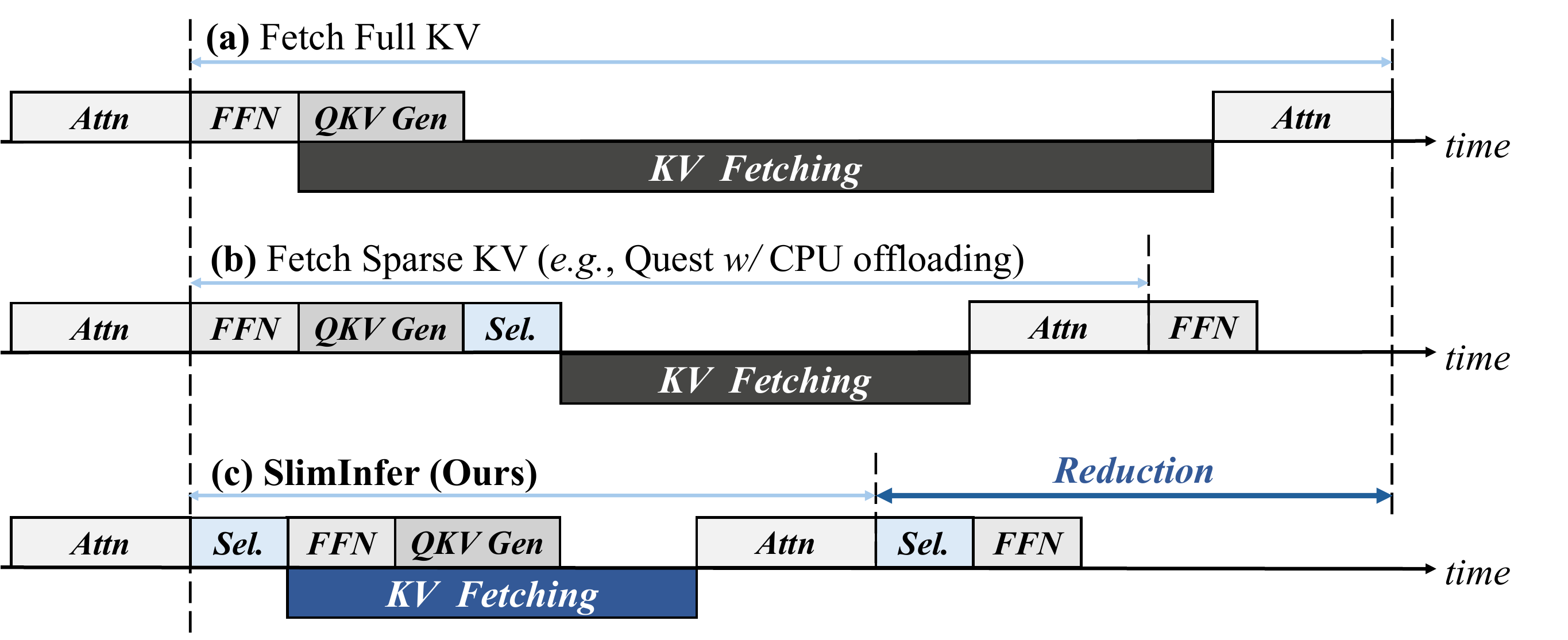} % Reduce the figure size so that it is slightly narrower than the column. Don't use precise values for figure width.This setup will avoid overfull boxes.
% TODO: Prefetch是否已经有了overlap的含义
\caption{SlimInfer reduces the latency of a layer (\emph{i.e.}, \textit{QKV Generation}+ \textit{Attention}+\textit{FFN}) by prefetching KV cache that is offloaded to CPU, overlapping KV cache fetching with computation. ``\textit{Sel.}'' means selecting tokens in KV cache (b) or hidden state (c) to prune.
}
\label{fig:kv_fetching}

\end{figure}

Managing the KV cache efficiently is a major challenge in long-context inference, especially when offloading the KV cache to the CPU for GPU memory savings. It introduces significant I/O costs by fetching offloaded KV cache from CPU to GPU during subsequent inference steps~\cite{lee2024infinigen}. To reduce this overhead, prior work introduces prefetching, a technique that overlaps KV cache transfer with computation to hide latency. However, enabling prefetching is non-trivial for token pruning that focuses on sparse attention. As shown in Figure~\ref{fig:kv_fetching} (b), Quest~\cite{tang2024quest} prunes tokens from the KV cache (including offloaded entries) based on current \textit{QKV} representations. After the pruning stage (\emph{i.e.}, \textit{Sel.}), the offloaded KV entries required for fetching are available.  Thus, it is impossible to overlap data transfer (\textit{KV Fetching}) with computation prior to \textit{Attention}. To allow prefetching, InfiniGen~\cite{lee2024infinigen} addresses this by rehearsing attention patterns using partial weights and offline SVD, while AttentionPredictor~\cite{yang2025attentionpredictor} trains a separate CNN to forecast future attention scores. Both approaches introduce additional computational and engineering overhead due to their speculative nature.

% However, existing token pruning studies prune tokens in the KV cache (including offloaded entries), which requires access to the current \textit{QKV}, and then fetch the offloaded parts of the pruned KV cache~\cite{tang2024quest}. As shown in Figure~\ref{fig:kv_fetching} (b), the pruned KV cache is only available after \textit{Sel.} and thus disables the potential of the I/O and computation overlap during \textit{KV Fetching}. 
% However, assessing the importance of a KV cache entry to enable prefetching in existing token pruning studies requires access to the current \textit{QKV}~\cite{tang2024quest}. As shown in Figure~\ref{fig:kv_fetching} (b), \textit{QKV} is only available after \textit{Sel.} and thus disables the potential of the I/O and computation overlap during \textit{KV Fetching}. To allow effective prefetching, InfiniGen~\cite{lee2024infinigen} addresses this by rehearsing attention patterns using partial weights and offline SVD, while AttentionPredictor~\cite{yang2025attentionpredictor} trains a separate CNN to forecast future attention scores. Both approaches introduce additional computational and engineering overhead due to their speculative nature.

\noindent\underline{\textit{Analysis.}} Notably, with the aforementioned hidden state pruning (Section~\ref{sec:inf}) applied following \textit{Attention} for a given layer, SlimInfer can eliminate the need for predictive mechanisms. As illustrated in Figure~\ref{fig:kv_fetching} (c), \textit{KV Fetching} can overlap with the computation of \textit{FFN} and \textit{QKV Generation} prior to the subsequent \textit{Attention}. Building upon this analysis, our framework can naturally achieve timely prefetching without any predictive or heuristic strategy.
%Notably, with the aforementioned hidden state pruning (Section~\ref{sec:inf}) applied following \textit{Attention} for a given layer, SlimInfer can eliminate the need for predictive mechanisms. 

\section{SlimInfer}
\begin{figure*}[t]
\centering
\includegraphics[width=1.8\columnwidth]{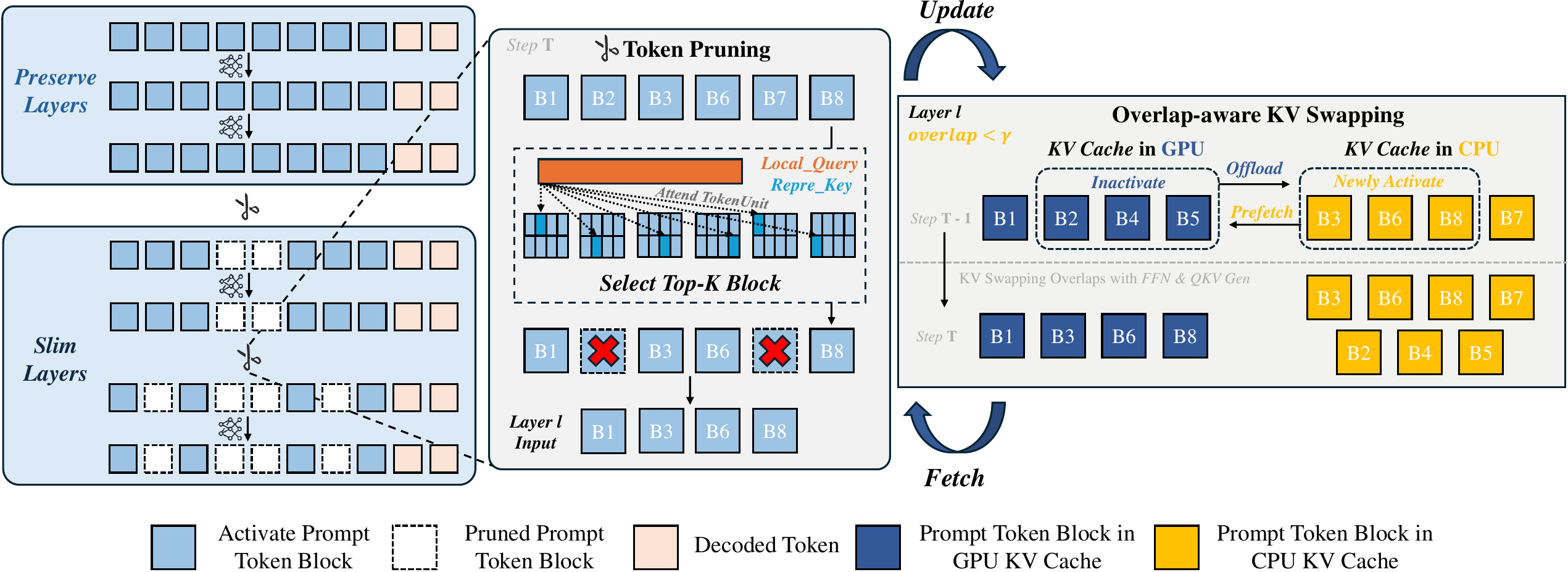} 
\caption{
Overview of the proposed \textbf{SlimInfer}. (\textit{i}) During inference, early \textit{Preserve Layers} retain all prompt blocks to support \textit{information diffusion} (Section~\ref{sec:inf}), while later \textit{Slim Layers} prune less relevant blocks to reduce computation (Section~\ref{sec:overview}). (\textit{ii}) Each block is divided into fine-grained \textit{Token Units} for accurate importance scoring (Section~\ref{sec:pruning}). (\textit{iii}) When $\texttt{overlap}<\gamma$ (Algorithm~\ref{alg:kv_swap}), SlimInfer triggers asynchronous KV cache swapping, which naturally overlaps data transfer (prefetching$+$offloading) with computation (Section~\ref{sec:free}). ``T'' denotes the current inference step.
}
\label{fig:overview}

\end{figure*}

% ------------------------------------------------------Overview--------------------------------------
\subsection{Framework Overview}
\label{sec:overview}
In this Section, we propose SlimInfer to accelerate long-context inference. It incorporates a dynamic block-wise hidden state pruning with a predictor-free KV cache prefetching strategy. Specifically, the prompt tokens are partitioned into fixed-size blocks, a common abstraction that aligns well with GPU-friendly batch operations and enables efficient memory access~\cite{tang2024quest, xiao2024infllm}. At any point during inference, a block is called \textit{active block} if it is deemed critical for ongoing computations. Only these active blocks participate in attention computation and have their KV entries stored in GPU memory. Additionally, our pruning mechanism is applied exclusively to prompt tokens. In contrast, all tokens generated as responses is fully retained to preserve fluency throughout generation. As shown in Figure~\ref{fig:overview}, inference is divided into two stages: 

\textit{\underline{Preserve layers.}} Motivated by our analysis of information diffusion, the early layers retain all tokens in the prompt. This ensures that critical semantic information has sufficient depth to propagate through the model before pruning. 

\textit{\underline{Slim layers.}} In the subsequent layers, SlimInfer dynamically prunes prompt blocks of hidden state across inference steps to reduce computation. This is guided by an accurate importance estimator that selects the top-$k$ most relevant blocks based on the recent decoding context. The pruning decisions (as demonstrated in Section~\ref{sec:pruning}) are made immediately after Attention computation and determine the \textit{active} block set for the next layer. This active block set is then propagated unchanged through subsequent layers until the next pruning operation.

The above hidden state pruning paradigm naturally enables efficient KV cache prefetching, as mentioned in Section~\ref{sec:pref}. Moreover, we adopt an overlap-aware design (see Section~\ref{sec:free}) for prefetching to avoid unnecessary data movement: Asynchronous KV cache prefetching and offloading are triggered only when the active block set changes significantly, enabling I/O to be overlapped with computation, thereby minimizing inference overhead.

% ------------------------------------------------------高效的渐进式prompt剪枝--------------------------------------
\subsection{Block-wise Prompt Token Pruning}
\label{sec:pruning}
Here we detail the specific pruning decision for \textit{Slim Layers}. Conventional approaches of block-level token pruning often estimate the contribution of each block to the current decoding context by compressing the entire block into a single vector~\cite{tang2024quest, yang2025attentionpredictor}, which may obscure fine-grained semantic information. To address this limitation, SlimInfer adopts a more expressive strategy by partitioning each prompt block of Key states into multiple smaller subsets, termed as \textit{token units}. This design captures finer-grained semantics within each block, enabling more accurate importance estimation without sacrificing block-level memory efficiency.

Specifically, each prompt block $B_j$ is divided into $M$ disjoint \textit{TokenUnits}, where each unit consists of a contiguous sequence of tokens within the block. For each token unit, a representative Key vector $k_{\text{rep}}(j, m)$ is computed by averaging the Key states of all tokens within that unit.
\begin{equation}
    k_{\text{rep}}(j, m) = \text{Mean} \left( \{ \text{key} \in \text{TokenUnit}_{j,m} \} \right).
    \label{eq:rep_key}
\end{equation}

To assess block importance, we construct a local window of the Query state, $q_l$, by averaging the Query vectors of the most recent $w$ tokens, drawn from the end of the prompt during the prefill phase or from decoded tokens during the decoding phase. For each representative Key vector $k_{\text{rep}}^{h}(j, m)$ in block $B_j$, we compute its similarity to $q_l$ via dot product on each attention head. The block-level importance score is then defined as:
\begin{equation}
    r_{\text{block}}(q_l, B_j) = \max_{m} \left\{ \frac{1}{H} \sum_{h=1}^{H} \left( q_l^{h} \cdot k_{\text{rep}}^{h}(j, m) \right) \right\},
    \label{eq:block_score}
\end{equation}
where $H$ denotes the number of attention heads, $m$ indexes token units within the block, and $h$ indexes attention heads.

In addition to this dynamic scoring, our pruning policy enforces the retention of structurally important blocks to maintain model stability. Specifically, the initial block of the prompt, which often acts as attention sinks~\cite{xiao2024efficientstreaminglanguagemodels}, is always preserved in the active set, regardless of its score. For all other blocks, the top-$k$ blocks with the highest importance scores are selected to form the candidate set, $B_{\text{candidate}}(t)$\footnote{$t$ denotes the current inference step.}, for subsequent computation. The KV cache of pruned blocks is not discarded but is offloaded to CPU memory, allowing future restoration when they become relevant again in subsequent decoding steps.

% ------------------------------------------------------预取KV cache机制--------------------------------------

\subsection{Predictor-Free KV Cache Prefetching}\label{sec:free}
\newcommand{\algcomment}[1]{\texttt{\textit{// #1}}}
\begin{algorithm}[tb]
\caption{Overlap-aware KV Swapping}
\label{alg:kv_swap}
\textbf{Input}: $B_{\text{active}}(t{-}1)$, $B_{\text{candidate}}(t)$, $B_{\text{Memory}}$, and threshold $\gamma$\\
\textbf{Output}: Updated $B_{\text{active}}(t)$
\begin{algorithmic}[1]
\STATE Compute overlap ratio:
\[
\texttt{overlap} \gets \frac{|B_{\text{candidate}}(t) \cap B_{\text{active}}(t{-}1)|}{|B_{\text{candidate}}(t)|}
\]
\IF{$\texttt{overlap} < \gamma$}
    \STATE $B_{\text{active}}(t) \gets B_{\text{candidate}}(t)$
    \STATE $B_{\text{offload}} \gets (B_{\text{active}}(t{-}1) \setminus B_{\text{active}}(t)) \setminus B_{\text{Memory}}$
    \STATE $B_{\text{load}} \gets (B_{\text{active}}(t) \setminus B_{\text{active}}(t{-}1))$
    \FOR{each block $B_j$ in $B_{\text{offload}}$ }
        \STATE \algcomment{Offloading} \\
        \STATE Move KV cache of $B_j$ from GPU to CPU
    \ENDFOR
    \FOR{each block $B_j$ in $B_{\text{load}}$ }
        \STATE \algcomment{Prefetching} \\
        \STATE Load KV cache of $B_j$ from CPU to GPU
    \ENDFOR
\ELSE
    \STATE $B_{\text{active}}(t) \gets B_{\text{active}}(t{-}1)$
\ENDIF
\end{algorithmic}
\end{algorithm}

To reduce GPU memory pressure, SlimInfer offloads the KV cache of inactive prompt blocks to the CPU. Under this scenario, SlimInfer naturally allows a predictor-free prefetching mechanism to reduce significant I/O costs that leverages its layer-wise hidden state pruning design (as demonstrated in Section~\ref{sec:pref}). Here, we further present an overlap-aware KV swapping (see Algorithm~\ref{alg:kv_swap}) to minimize unnecessary data transfer for prefetching as follows.

% paragraph : prefetching trigger mechanism
At each inference step $t\,(t>1)$, SlimInfer maintains an active block set\footnote{The definition of active block can be found in Section~\ref{sec:overview}.}, $B_{\text{active}}(t)$, whose corresponding KV cache entries are stored in GPU memory for fast access. To minimize unnecessary data movement, a swap operation (\emph{i.e.}, offloading $+$ prefetching) is only enabled when the composition of this set needs to change significantly. Specifically, SlimInfer first establishes the candidate activation set, $B_{\text{candidate}}(t)$ (see Section~\ref{sec:pruning}), calculated based on importance scores. SlimInfer then computes the overlap ratio between this candidate set and the previous active block set, $B_{\text{active}}(t-1)$. If the ratio falls below a predefined threshold $\gamma$, a swap operation is triggered. Otherwise, $B_{\text{active}}(t-1)$ is directly reused as $B_{\text{active}}(t)$, which neglects KV cache prefetching and incurs negligible performance drops (see Appendix). This design prioritizes inference efficiency to reduce data transfer overhead.
% The size of this set (\textit{i.e.}, the number of active blocks), is fixed for each layer throughout inference, ensuring a constant pruning ratio.
                                                             
% Asynchronous Swapping Operation
The swap operation, as detailed in Algorithm~\ref{alg:kv_swap}, involves asynchronous prefetching: (\textit{i}) KV entries for newly required blocks ($B_{\text{load}}$) are transferred from the CPU to the GPU. (\textit{ii}) Entries for unneeded blocks ($B_{\text{offload}}$) that are not yet in the CPU memory pool are offloaded to the CPU; those already reside in CPU (\emph{i.e.}, corresponding to blocks $B_{\text{Memory}}$), their GPU memory is immediately released for newly prefetched entries. To maximize efficiency and hide I/O latency, the offloading and prefetching processes are executed on a separate CUDA stream. As illustrated in Figure~\ref{fig:kv_fetching}, this swap overlaps with the subsequent \textit{FFN} and \textit{QKV Generation}.

\section{Experiments}
\begin{table*}[t]
\centering
\setlength{\tabcolsep}{2pt}
{\fontsize{9}{11}\selectfont
\begin{tabular}{lcccccccccccccccccc}
\toprule
\multicolumn{1}{c}{\multirow{4}{*}{Method}}     & \multicolumn{2}{c}{\makecell{Single-Doc. \\QA}}                      & \multicolumn{3}{c}{\makecell{Multi-Doc.\\ QA}}                                                     & \multicolumn{4}{c}{Summarization}                                                                                     & \multicolumn{3}{c}{\makecell{Few-shot\\ Learning}}                                                       & \multicolumn{3}{c}{\makecell{Synthetic\\ Task}}                      & \multicolumn{2}{c}{\makecell{Code\\ Completion}}                    & \multicolumn{1}{c}{\multirow{4}{*}{Avg. (\%)}}                        \\

\cmidrule(lr){2-3}\cmidrule{4-6}\cmidrule(lr){7-10}\cmidrule(lr){11-13}\cmidrule{14-16}\cmidrule(lr){17-18}
 & \multicolumn{1}{c}{\rotatebox{90}{Qasper}} & \multicolumn{1}{c}{\rotatebox{90}{MQA}} & \multicolumn{1}{c}{\rotatebox{90}{HPQA}} & \multicolumn{1}{c}{\rotatebox{90}{2WiKi}} & \multicolumn{1}{c}{\rotatebox{90}{MuSiQue}} & \multicolumn{1}{c}{\rotatebox{90}{GovRep}} & \multicolumn{1}{c}{\rotatebox{90}{QMSum}} & \multicolumn{1}{c}{\rotatebox{90}{MNews}} & \multicolumn{1}{c}{\rotatebox{90}{VCSum}} & \multicolumn{1}{c}{\rotatebox{90}{TREC}} & \multicolumn{1}{c}{\rotatebox{90}{TQA}} & \multicolumn{1}{c}{\rotatebox{90}{SAMSum}} & \multicolumn{1}{c}{\rotatebox{90}{LSHT}} & \multicolumn{1}{c}{\rotatebox{90}{Count}} & \multicolumn{1}{c}{\rotatebox{90}{PassR}} & \multicolumn{1}{c}{\rotatebox{90}{LCC}} & \multicolumn{1}{c}{\rotatebox{90}{RepB-p}} &  \\ \midrule
\multicolumn{19}{c}{\textit{LLaMA3.1-8B-Instruct}}\\
\midrule
Full KV     & 45.82           & 55.05          & 55.50          & 44.28          & 30.78            & 35.21           & 25.49          & 27.23          & 17.17          & 72.50          & 91.65          & 43.92           & 46.00          & 7.43           & 99.50           & 63.12          & 56.74           & 48.08            \\
\midrule
LazyLLM     & \textbf{46.39}  & 51.28          & 54.52          & 43.42          & 28.86            & 34.57           & 25.41          & 27.05          & \underline{17.30}    & 70.50          & 91.00          & 43.64           & \textbf{46.00} & \textbf{7.94}  & \textbf{99.50}  & 59.44          & 56.12           & \underline{47.23}      \\
MInference  & 44.29           & 52.53          & 52.00          & \underline{44.10}    & 25.72            & \textbf{35.09}  & \underline{25.47}    & \textbf{27.21} & \textbf{17.53} & \textbf{72.00} & \underline{91.18}    & \underline{43.73}     & \textbf{46.00} & 3.25           & 97.00           & \textbf{64.87} & \underline{60.00}     & 47.17            \\
FlexPrefill & 44.55           & \textbf{55.56} & \underline{54.56}    & 43.43          & \underline{30.07}      & 34.64           & \textbf{25.83} & 27.05          & 16.97          & 70.50          & 89.81          & 43.18           & 41.00          & 2.59           & 82.00           & \underline{64.67}    & \textbf{62.06}  & 46.38            \\
SlimInfer   & \underline{45.19}     & \underline{53.82}    & \textbf{55.14} & \textbf{44.37} & \textbf{30.95}   & \underline{34.99}     & 24.77          & \underline{27.10}    & 16.81          & \underline{71.00}    & \textbf{91.65} & \textbf{44.36}  & 45.50          & \underline{6.30}     & \underline{98.50}     & 63.65          & 55.95           & \textbf{47.65}   \\
\midrule
\multicolumn{19}{c}{\textit{Qwen2.5-7B-Instruct}}\\
\midrule
% FULL 行
Full KV     & 43.92           & 52.76          & 57.97          & 46.56          & 30.16            & 31.78           & 23.36          & 24.30          & 16.05          & 72.50          & 88.64          & 45.64           & 43.00          & 8.00           & 100.00          & 60.44          & 66.84           & 47.76            \\
\midrule
LazyLLM     & 39.79           & 45.71          & 53.30          & 42.58          & 28.94            & 31.16           & 23.08          & 23.28          & 15.61          & 66.50          & 87.67          & 45.31           & \underline{42.25}    & 6.59           & \textbf{100.00} & 57.46          & 63.89           & 45.48            \\
MInference  & \textbf{44.02}  & \textbf{52.86} & \textbf{58.25} & \underline{46.17}    & \textbf{29.85}   & \underline{31.78}     & \underline{23.27}    & 23.88          & 15.84          & \underline{71.50}    & \textbf{89.09} & \underline{45.89}     & 41.60          & \underline{8.00}     & 92.00           & \textbf{61.33} & \textbf{67.98}  & \underline{47.25}      \\
FlexPrefill & 41.65           & 51.92          & 55.29          & 41.65          & \underline{29.69}      & 31.71           & \underline{23.27}    & \underline{24.05}    & \underline{15.91}    & 70.50          & 88.22          & \textbf{46.45}  & 36.50          & 2.00           & 75.00           & \underline{61.10}    & 63.38           & 44.61            \\
SlimInfer   & \underline{43.74}     & \underline{52.31}    & \underline{56.94}    & \textbf{46.62} & 27.25            & \textbf{31.85}  & \textbf{23.40} & \textbf{24.26} & \textbf{16.09} & \textbf{72.00} & \underline{89.01}    & 45.52           & \textbf{43.00} & \textbf{8.50}  & \underline{99.00}     & 60.21          & \underline{65.71}     & \textbf{47.38}   \\ 
\bottomrule
\end{tabular}
}
\caption{Performance comparison on LongBench~\cite{bai2024longbenchbilingualmultitaskbenchmark}. The best and second results are in \textbf{bold} and \underline{underlined}.}
\label{tab:longbench}

\end{table*}

\subsection{Settings}\label{sec:settings}

\subsubsection{Models}
% TOOD ： 补充引用
The experiments are conducted using LLaMA-3.1-8B-Instruct (LLaMA-3.1)~\cite{grattafiori2024llama3herdmodels}  and Qwen2.5-7B-Instruct (Qwen-2.5)~\cite{qwen2025qwen25technicalreport} to evaluate the effectiveness of our method in larger-scale LLMs. Both models support context lengths of 128k. 

%The input prompts are constructed using the default chat template.

\subsubsection{Implementation Details}
Our framework is built on LazyLLM~\cite{fu2024lazyllmdynamictokenpruning} and is implemented in PyTorch. For the inference pipeline, we integrate SlimInfer into the Transformers~\cite{wolf2020huggingfacestransformers} library by replacing the default self-attention module to support efficient block-wise token pruning and asynchronous KV cache management. Unless otherwise noted, we use a block size of 64, a token unit size of 8, a KV swap threshold $\gamma = 0.9$, and a local query window of 4. Pruning is applied at layers 10, 20, and 30 for LLaMA3.1, retaining 8k, 4k, and 2k tokens respectively; and at layers 9, 18, and 26 for Qwen2.5, retaining 12k, 6k, and 4k tokens. All accuracy experiments are conducted on an NVIDIA H200 GPU, while efficiency evaluations are run on a single NVIDIA RTX 4090 GPU (24GB) to simulate typical edge deployment. 

% All experiments use BF16 precision with greedy decoding.

\subsubsection{Baselines}
To evaluate the effectiveness of \textbf{SlimInfer}, we compare it with FlashAttention2 (Full KV) \cite{dao2023flashattention2fasterattentionbetter} and 3 token pruning approaches for long-context processing: MInference \cite{jiang2024minference10acceleratingprefilling}, FlexPrefill \cite{lai2025flexprefillcontextawaresparseattention}, and LazyLLM \cite{fu2024lazyllmdynamictokenpruning}. FlashAttention2 serves as the dense attention baseline, while the others adopt sparse attention or memory management to improve efficiency. All results are based on public implementations. To ensure a fair comparison, LazyLLM applies pruning at the same layers as SlimInfer, retaining 50\% of tokens at each pruning layer. For FlexPrefill, we use $\gamma = 0.95$ for both LLaMA-3.1 and Qwen-2.5, consistent with its recommended configuration. For MInference, we follow its official codebase and select the sparse attention pattern for each head accordingly.

\subsection{Accuracy Evaluation}

% TODO：补充引用
Following common practice~\cite{zhang2025spargeattention, li2024snapkvllmknowslooking, zhang2025pqcache}, we adopt the LongBench~\cite{bai2024longbenchbilingualmultitaskbenchmark} to evaluate the generation quality of our method under long-context understanding settings. LongBench includes a wide range of tasks such as single-document and multi-document QA, summarization, few-shot learning, synthetic tasks, and code completion. Each task is evaluated using task-specific metrics such as accuracy, F1-score, and Rouge-L, where higher scores indicate better performance.

As shown in Table~\ref{tab:longbench}, SlimInfer consistently achieves the highest average accuracy across both LLaMA3.1-8B-Instruct and Qwen2.5-7B-Instruct models. Beyond its strong overall performance, SlimInfer exhibits consistent and robust accuracy across diverse task categories, matching or surpassing other baselines on most benchmarks. These results underscore its broad generalization capability across different model architectures.

% Table~\ref{tab:longbench} reports the accuracy of different methods on LLaMA3.1-8B-Instruct and Qwen2.5-7B-Instruct across LongBench tasks. SlimInfer achieves the highest average accuracy across both models, consistently matching or surpassing other efficient baselines.

\subsection{Efficiency Evaluation}
\label{sec:efficiency_eval}
\begin{figure}[!ht]
\centering
\includegraphics[width=\columnwidth]{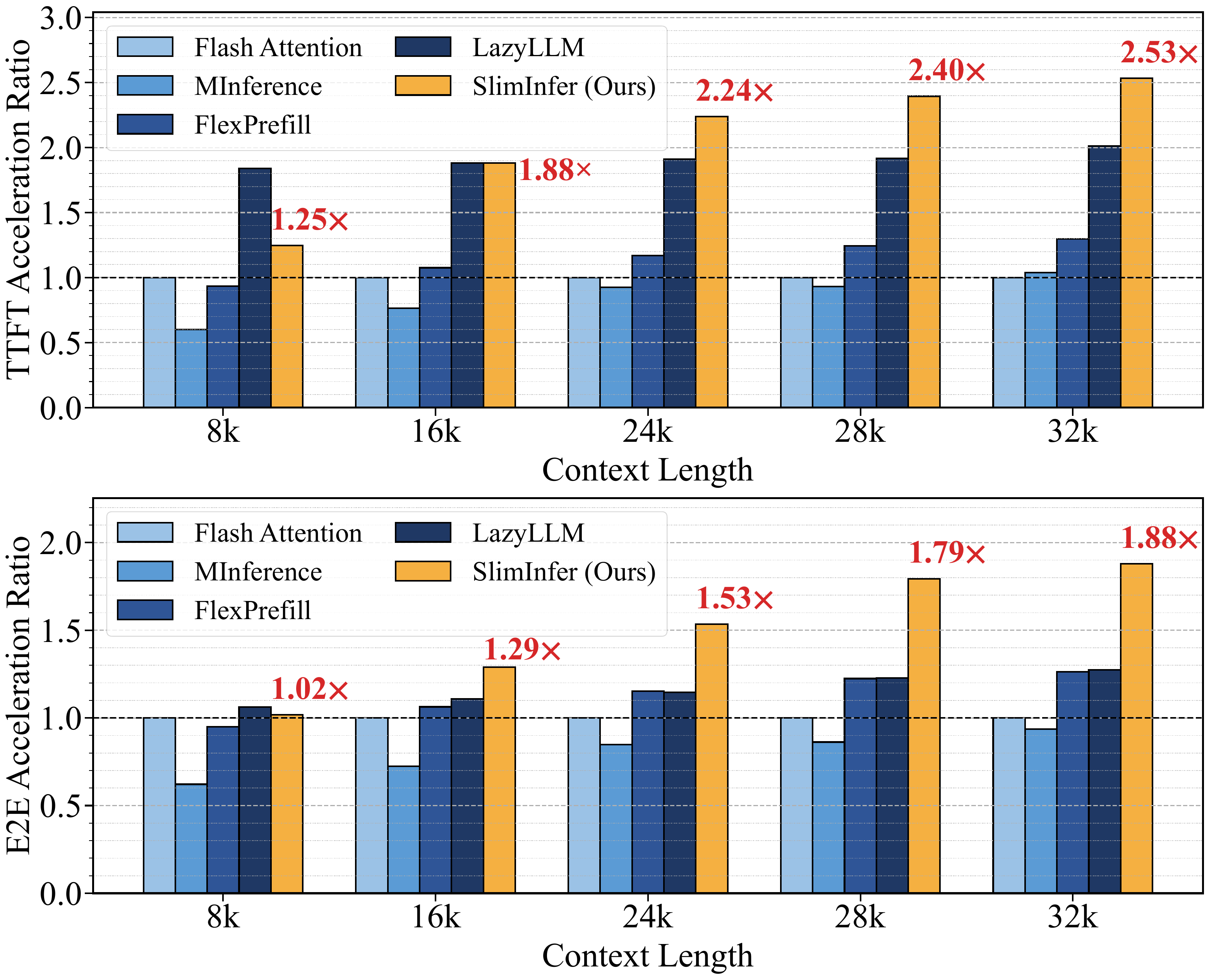} 
\caption{Inference efficiency comparison for LLaMA3.1-8B-Instruct~\cite{grattafiori2024llama3herdmodels}. (\textbf{Upper}) TTFT and (\textbf{Lower}) E2E latency acceleration ratio \emph{vs.} context length. SlimInfer far outperforms the baselines at long context lengths ($\geq$24k) and remains on par with them in other cases.}
\label{fig:llam3.1-efficiency}
\end{figure}

\subsubsection{Latency Profiling} We benchmark the inference latency across various methods with a single input sequence. All experiments are conducted on an RTX 4090 GPU using LLaMA-3.1-8B-Instruct~\cite{grattafiori2024llama3herdmodels}. To assess how latency scales with input length, we use 5 truncated versions of a 32k token sequence sampled from LongBench~\cite{bai2024longbenchbilingualmultitaskbenchmark}. We report two metrics: (1) \textit{Time-to-First-Token (TTFT)} latency, and (2) \textit{End-to-End (E2E)} latency for decoding 16 tokens. In Figure~\ref{fig:llam3.1-efficiency}, we present the acceleration ratios of various inference baselines relative to the FlashAttention2~\cite{dao2023flashattention2fasterattentionbetter} baseline. Across all input lengths, our method consistently achieves significant speedups in both TTFT and E2E latency. In particular, our SlimiInfer shows an increasing acceleration trend for TTFT as context length grows, highlighting the advantage of our sparse prefill design in long-context scenarios. Compared to other baselines, our method achieves the highest TTFT speedup (up to $\mathbf{2.53\times}$) and E2E speedup (up to $\mathbf{1.88\times}$) at 32k input length. These results further validate the superiority of our design in reducing both prompt prefilling and decoding latency. Comparison for Qwen2.5-7B-Instruct~\cite{qwen2025qwen25technicalreport} can be found in Appendix.

\begin{figure}[!ht]
\centering
\includegraphics[width=1.0\columnwidth]{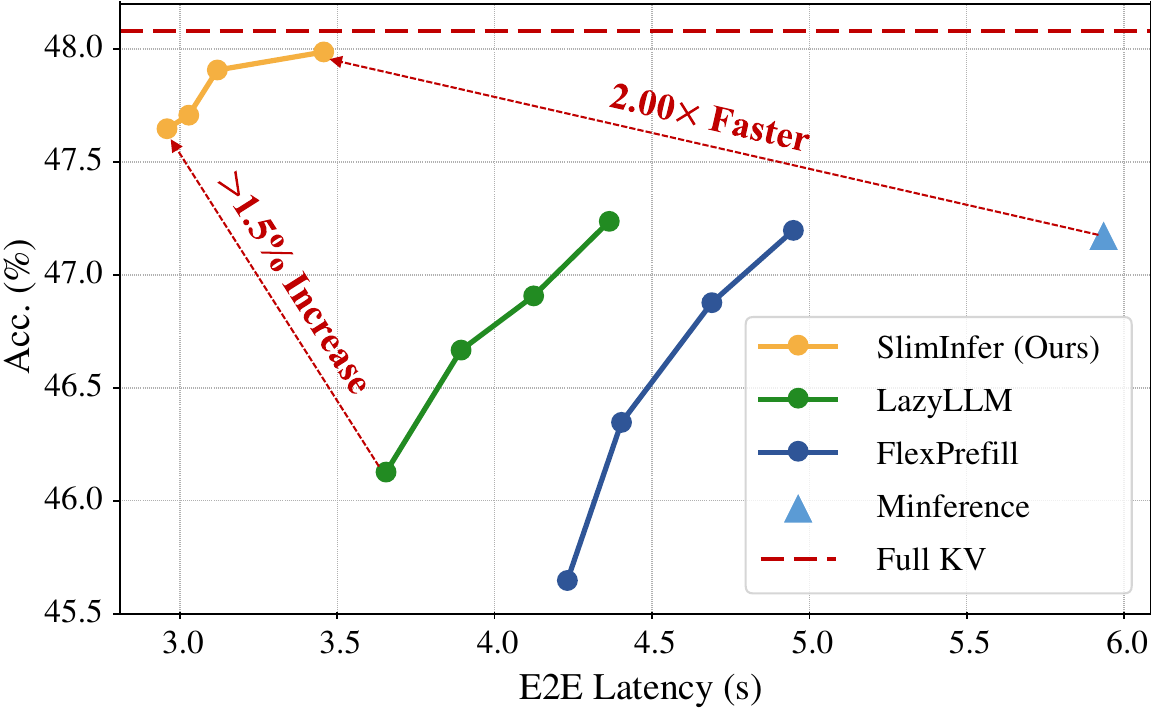} 
\caption{Accuracy \emph{vs.} E2E latency for LLaMA3.1-8B-Instruct~\cite{grattafiori2024llama3herdmodels} on LongBench~\cite{bai2024longbenchbilingualmultitaskbenchmark} with 32k context. SlimInfer achieves a markedly superior trade‑off, delivering near-lossless accuracy drops with substantially lower latency than other methods.}
\label{fig:llam3.1-pareto}
\end{figure}

\subsubsection{Accuracy \emph{vs.} Efficiency}
Our dynamic pruning strategy enables flexible trade-offs between inference efficiency and model accuracy. In Figure~\ref{fig:llam3.1-pareto}, we compare end-to-end latency and LongBench~\cite{bai2024longbenchbilingualmultitaskbenchmark} accuracy across different baselines. The results show that SlimInfer establishes a strong Pareto frontier: It achieves accuracy close to the full KV baseline while substantially reducing latency. Compared to existing methods, SlimInfer offers a more favorable balance between quality and efficiency.

\subsubsection{Memory Efficiency}

\begin{table}[!ht]
\centering
{\fontsize{9}{11}\selectfont
\begin{tabular}{lccccc}
\toprule
Method & 8k & 16k & 24k & 28k & 32k \\
\midrule
Full KV (Baseline)   & 1.00 & 2.00 & 3.00 & 3.50 & 4.00 \\
SlimInfer (Ours)     & 0.80 & 1.11 & 1.42 & 1.58 & 1.73 \\
Memory Saving (\%)   & \textbf{20.3} & \textbf{44.5} & \textbf{52.6} & \textbf{54.9} & \textbf{56.6} \\
\bottomrule
\end{tabular}
}
\caption{Prompt KV cache memory consumption (GB) on LLaMA-3.1-8B-Instruct across different input lengths.}
\label{tab:memory_efficiency}
\end{table}

In addition to latency, we evaluate SlimInfer's GPU memory footprint against other representative methods. FlexPrefill~\cite{lai2025flexprefillcontextawaresparseattention} and MInference~\cite{jiang2024minference10acceleratingprefilling} optimize computation but retain the full KV cache throughout all layers, resulting in no memory savings. LazyLLM~\cite{fu2024lazyllmdynamictokenpruning} applies dynamic pruning but overlooks KV cache offloading, missing an opportunity to reduce substantial GPU memory overhead. In contrast, SlimInfer combines a dynamic pruning strategy with offloading for KV pairs from inactive blocks to CPU memory. Therefore, SlimInfer effectively limits GPU memory usage throughout inference. As shown in Table~\ref{tab:memory_efficiency}, this design yields 20.3$-$56.6\% reductions in prompt KV cache memory.

\subsection{Ablation Study}
We use LLaMA3.1‑8B‑Instruct~\cite{grattafiori2024llama3herdmodels} here. 
% The default settings are given in Section~\ref{sec:settings}.
\subsubsection{Balancing Pruning Depth and Token Retention}
We vary the pruning start layer while keeping the total number of retained tokens constant. This allows us to examine how the pruning position affects model performance across tasks. Specifically, pruning occurs only once at the start layer.

% As shown in Figure~\ref{fig:pruning_layer_impact}, all three tasks exhibit a clear non-linear trend: performance improves as pruning is delayed, peaking around the middle layers. However, further delaying pruning leads to a sharp accuracy drop for MQA and Qasper, while Pass Retrieval remains largely stable. This degradation is caused by a substantial drop in the number of retained tokens needed to satisfy the sparsity constraint when pruning is applied at later layers. Early pruning disrupts the model’s ability to perform information compression, whereas overly late pruning leads to insufficient token capacity for downstream reasoning. QA tasks such as MQA and Qasper are particularly sensitive to this constraint, while retrieval tasks like PassR exhibit greater robustness. These findings suggest that pruning should avoid disrupting early information diffusion and ensure enough tokens are retained in later layers to support downstream reasoning.

As shown in Figure~\ref{fig:pruning_layer_impact}, all three tasks exhibit a non-linear pattern: Accuracy improves as pruning is delayed to the middle layers. However, for MQA and Qasper, further delaying pruning causes a sharp accuracy drop, while PassR remains largely stable. This could result from too few tokens being retained under the sparsity constraint at later layers. Early pruning hinders \textit{information diffusion}, whereas late pruning restricts token capacity for downstream reasoning. This underscores the need to balance early information preservation with sufficient late-layer token availability.

\begin{figure}[t]
\centering
\includegraphics[width=1.0\columnwidth]{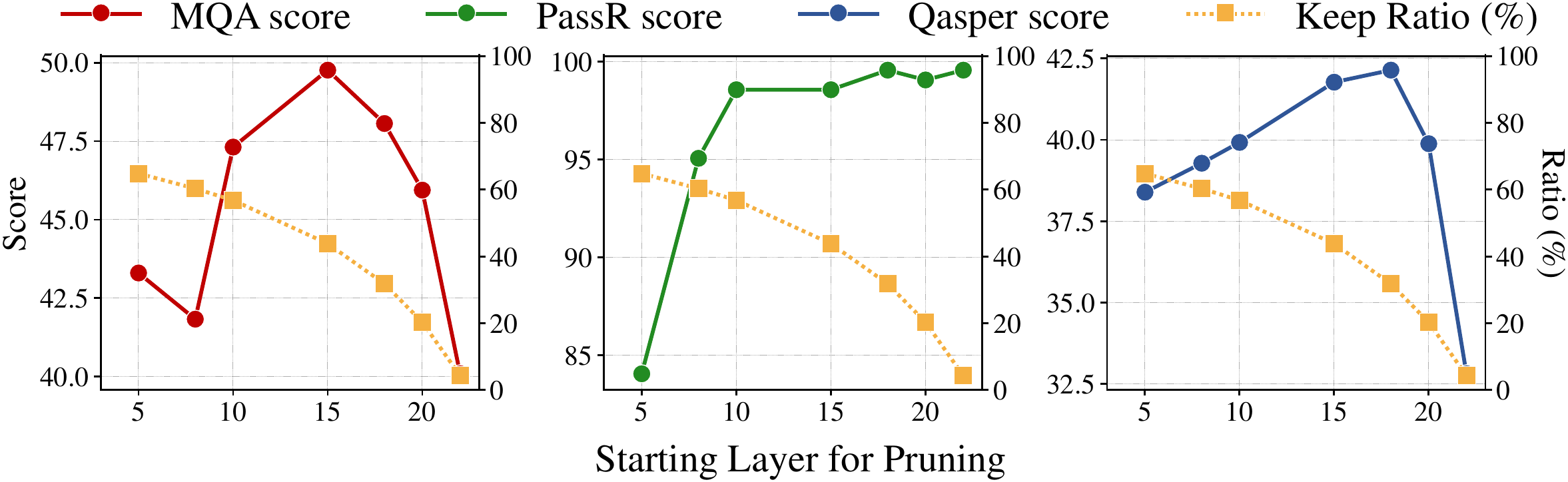} 
\caption{Impact of pruning start layer on task performance under fixed overall sparsity.}
\label{fig:pruning_layer_impact}
\end{figure}

\begin{table}[!ht]
\centering
{\fontsize{9}{11}\selectfont
\begin{tabular}{lccccc}
\toprule
Method & MuSiQue & PassR & HPQA & Avg. Score \\
\midrule
Avg-Pooling & 30.52 & 95.00 & 55.03 & 47.45 \\
Max-Pooling & 29.77 & 98.00 & 54.31 & 47.44 \\
SlimInfer & \textbf{30.95} & \textbf{98.50} & \textbf{55.14} & \textbf{47.65} \\
\bottomrule
\end{tabular}%
}
\caption{Ablation study on block importance scoring methods on LongBench~\cite{bai2024longbenchbilingualmultitaskbenchmark}.}
\label{tab:ablation_scoring}
\end{table}

\subsubsection{Block Importance Scoring Algorithm}
To assess the effectiveness of our block-wise pruning algorithm, we ablate the block importance scoring strategy. We compare our Token Unit–based method, which partitions each block into finer-grained token units, against two baselines: Avg-Pooling (average of token key states) and Max-Pooling (element-wise maximum). All other SlimInfer settings are
 kept constant. As shown in Table~\ref{tab:ablation_scoring}, our approach consistently outperforms the baselines across representative tasks, achieving the highest average score. This highlights the advantage of finer-grained representations in capturing semantic importance for more effective pruning.

\begin{table}[!ht]
\centering
\setlength{\tabcolsep}{1mm}
{\fontsize{9}{11}\selectfont
\begin{tabular}{lccccc}
\toprule
Method & 8k & 16k & 24k & 28k & 32k \\
\midrule
Full KV (Baseline) & 1.00$\times$ & 1.00$\times$ & 1.00$\times$ & 1.00$\times$ & 1.00$\times$ \\
Ours (w/o async KV)  & 1.00$\times$ & 1.18$\times$ & 1.37$\times$ & 1.48$\times$ & 1.60$\times$ \\
Ours (w/ async KV) & \textbf{1.02$\times$} & \textbf{1.29$\times$} & \textbf{1.53$\times$} & \textbf{1.79$\times$} & \textbf{1.88$\times$} \\
\bottomrule
\end{tabular}
}
\caption{End-to-end inference speedup across input lengths for LLaMA3.1-8B-Instruct.}
\label{tab:async_kv_speedup}
\end{table}

\subsubsection{Overlapping Operations for Latency Reduction}
To evaluate the impact of asynchronous KV cache management, we compare end-to-end inference latency with and without this optimization. As shown in Table~\ref{tab:async_kv_speedup}, our method achieves consistent speedups over the FlashAttention baseline across input lengths. At 32k context length, SlimInfer reaches a $1.60\times$ speedup without async KV and further improves to $1.88\times$ with it. The gains increase with input length, demonstrating the effectiveness of overlapping computation and data transfer for long-context inference.

\section{Conclusion}
We introduce SlimInfer, a framework that accelerates long-context LLM inference through dynamic block-wise token pruning for the hidden state. To preserve essential context, SlimInfer adopts fine-grained importance evaluation to guide accurate and efficient pruning. This deterministic design further supports a predictor-free asynchronous KV cache manager that effectively hides I/O latency. Extensive experiments demonstrate that SlimInfer significantly improves both Time-To-First-Token and end-to-end latency, without compromising performance.

%------------------------------------正文结束------------------------------------

\section*{Acknowledgments}
This work is supported in part by the National Key R\&D Program of China (Grant No. 2023YFB4503704 and 2024YFB4505601), the National Natural Science Foundation of China (Grant No. 62572036), and the Beijing Natural Science Foundation (Grant No. L243031).

\bibliography{aaai2026}

% \newpage

\fi

% \appendix
\begin{center}
    \Large{\textbf{Appendix}}
\end{center}

\ifappendixonly
\setcounter{page}{1}
% \linenumbers
% \setcounter{linenumber}{1}
\else
\fi
\setcounter{section}{0}
\setcounter{equation}{0}
\setcounter{figure}{0}
\renewcommand\thesection{\Alph{section}}
\renewcommand\thefigure{\Alph{figure}}
\renewcommand\thetable{\Alph{table}}
\renewcommand{\theequation}{\Alph{equation}}

\begin{table*}[ht!]
\centering
\setlength{\tabcolsep}{2.5pt} % Slightly adjusted column separation for better fit
\fontsize{8}{10}\selectfont % Adjusted font size for better fit
\begin{tabular}{l@{\hskip 8pt}lcccccccccccccccccc@{\hskip 8pt}c}
\toprule
\multicolumn{2}{c}{\multirow{2}{*}{Method}} & \multicolumn{2}{c}{\makecell{Single-Doc. \\QA}} & \multicolumn{3}{c}{\makecell{Multi-Doc.\\ QA}} & \multicolumn{4}{c}{Summarization} & \multicolumn{3}{c}{\makecell{Few-shot\\ Learning}} & \multicolumn{3}{c}{\makecell{Synthetic\\ Task}} & \multicolumn{2}{c}{\makecell{Code\\ Completion}} & \multirow{2}{*}{\makecell{Avg.\\(\%)}} & \multirow{2}{*}{\makecell{E2E \\ (s)}} \\
\cmidrule(lr){3-4}\cmidrule(lr){5-7}\cmidrule(lr){8-11}\cmidrule(lr){12-14}\cmidrule(lr){15-17}\cmidrule(lr){18-19}
& & \rotatebox{90}{Qasper} & \rotatebox{90}{MQA} & \rotatebox{90}{HPQA} & \rotatebox{90}{2WiKi} & \rotatebox{90}{MuSiQue} & \rotatebox{90}{GovRep} & \rotatebox{90}{QMSum} & \rotatebox{90}{MNews} & \rotatebox{90}{VCSum} & \rotatebox{90}{TREC} & \rotatebox{90}{TQA} & \rotatebox{90}{SAMSum} & \rotatebox{90}{LSHT} & \rotatebox{90}{Count} & \rotatebox{90}{PassR} & \rotatebox{90}{LCC} & \rotatebox{90}{RepB-p} & & \\ 
\midrule
\multicolumn{21}{c}{\textit{LLaMA3.1-8B-Instruct}} \\
\midrule
\multicolumn{2}{l}{Full KV} & 45.82 & 55.05 & 55.50 & 44.28 & 30.78 & 35.21 & 25.49 & 27.23 & 17.17 & 72.50 & 91.65 & 43.92 & 46.00 & 7.43 & 99.50 & 63.12 & 56.74 & 48.08 & 5.561 \\
\midrule
\multicolumn{2}{l}{MInference} & 44.29 & 52.53 & 52.00 & 44.10 & 25.72 & 35.09 & 25.47 & 27.21 & 17.53 & 72.00 & 91.18 & 43.73 & 46.00 & 3.25 & 97.00 & 64.87 & 60.00 & 47.17 & 5.935 \\
\midrule
\multirow{4}{*}{FlexPrefill} & $\gamma=0.99$ & 44.06 & 55.64 & 55.07 & 44.91 & 32.31 & 35.01 & 25.22 & 27.13 & 17.39 & 72.00 & 91.65 & 43.93 & 45.50 & 3.19 & 85.50 & 64.29 & 59.40 & 47.19 & 4.949 \\
& $\gamma=0.98$ & 44.83 & 57.31 & 55.93 & 42.14 & 30.95 & 34.85 & 24.97 & 27.13 & 17.27 & 69.50 & 91.64 & 44.07 & 45.50 & 4.31 & 81.00 & 64.30 & 61.06 & 46.87 & 4.690 \\
& $\gamma=0.95$ & 44.55 & 55.56 & 54.56 & 43.43 & 30.07 & 34.64 & 25.83 & 27.05 & 16.97 & 70.50 & 89.81 & 43.18 & 41.00 & 2.59 & 82.00 & 64.67 & 62.06 & 46.38 & 4.403 \\
& $\gamma=0.90$ & 43.64 & 54.56 & 55.56 & 35.74 & 26.05 & 34.53 & 25.07 & 27.11 & 17.13 & 69.00 & 91.04 & 42.93 & 40.00 & 2.16 & 82.50 & 65.56 & 63.33 & 45.64 & 4.231 \\
\midrule
\multirow{4}{*}{LazyLLM} & 50\% & 46.39 & 51.28 & 54.52 & 43.42 & 28.86 & 34.57 & 25.41 & 27.05 & 17.30 & 70.50 & 91.00 & 43.64 & 46.00 & 7.94 & 99.50 & 59.44 & 56.12 & 47.23 & 4.364 \\
& 45\% & 46.78 & 53.69 & 49.34 & 42.94 & 28.29 & 34.39 & 25.35 & 26.63 & 17.26 & 70.00 & 91.52 & 43.94 & 46.00 & 7.47 & 99.50 & 58.47 & 55.78 & 46.90 & 4.124 \\
& 40\% & 44.31 & 55.51 & 49.12 & 42.61 & 30.14 & 33.97 & 25.33 & 26.67 & 17.60 & 69.00 & 91.23 & 43.86 & 45.50 & 7.24 & 99.50 & 56.28 & 55.30 & 46.66 & 3.893 \\
& 35\% & 42.43 & 52.76 & 47.41 & 43.14 & 30.74 & 33.98 & 24.71 & 26.66 & 17.05 & 69.00 & 91.23 & 43.71 & 46.00 & 6.33 & 99.50 & 55.17 & 54.14 & 46.12 & 3.655 \\
\midrule
\multirow{4}{*}{\makecell{SlimInfer\\(Ours)}} & (16, 8, 4) & 45.62 & 55.93 & 54.86 & 43.43 & 30.00 & 34.87 & 24.88 & 27.20 & 17.58 & 72.50 & 91.47 & 44.31 & 46.50 & 6.96 & 99.50 & 63.33 & 56.73 & 47.98 & 3.457 \\
& (12, 6, 4) & 45.58 & 53.55 & 54.76 & 45.25 & 30.72 & 35.15 & 25.15 & 27.20 & 17.42 & 72.50 & 91.48 & 44.27 & 45.50 & 6.94 & 99.00 & 63.46 & 56.41 & 47.90 & 3.119 \\
& (10, 5, 3) & 45.71 & 53.95 & 54.45 & 43.51 & 29.78 & 34.97 & 25.09 & 27.24 & 17.12 & 71.50 & 91.48 & 44.33 & 46.00 & 6.88 & 99.00 & 63.46 & 56.41 & 47.70 & 3.028 \\
& (8, 4, 2) & 45.19 & 53.82 & 55.14 & 44.37 & 30.95 & 34.99 & 24.77 & 27.10 & 16.81 & 71.00 & 91.65 & 44.36 & 45.50 & 6.30 & 98.50 & 63.65 & 55.95 & 47.65 & 2.959 \\
\bottomrule
\end{tabular}
\caption{Detailed performance and latency comparison on LongBench~\cite{bai2024longbenchbilingualmultitaskbenchmark} for LLaMA3.1-8B-Instruct~\cite{grattafiori2024llama3herdmodels}. We report results for baselines and various configurations of our method, SlimInfer. The numbers in parentheses for SlimInfer, \emph{e.g.}, (16,8,4), denote the number of retained prompt tokens (in units of k, where $1\text{k} = 1024$) at each of the three pruning layers, respectively. The E2E Latency column shows the time in seconds to generate 16 tokens with a 32k input.}
\label{tab:longbench_appendix_llama}
\end{table*}

\section{Additional Accuracy Experiments}
\subsection{Needle-in-a-Haystack Result}

\begin{figure}[h!]
\centering
\includegraphics[width=1.0\columnwidth]{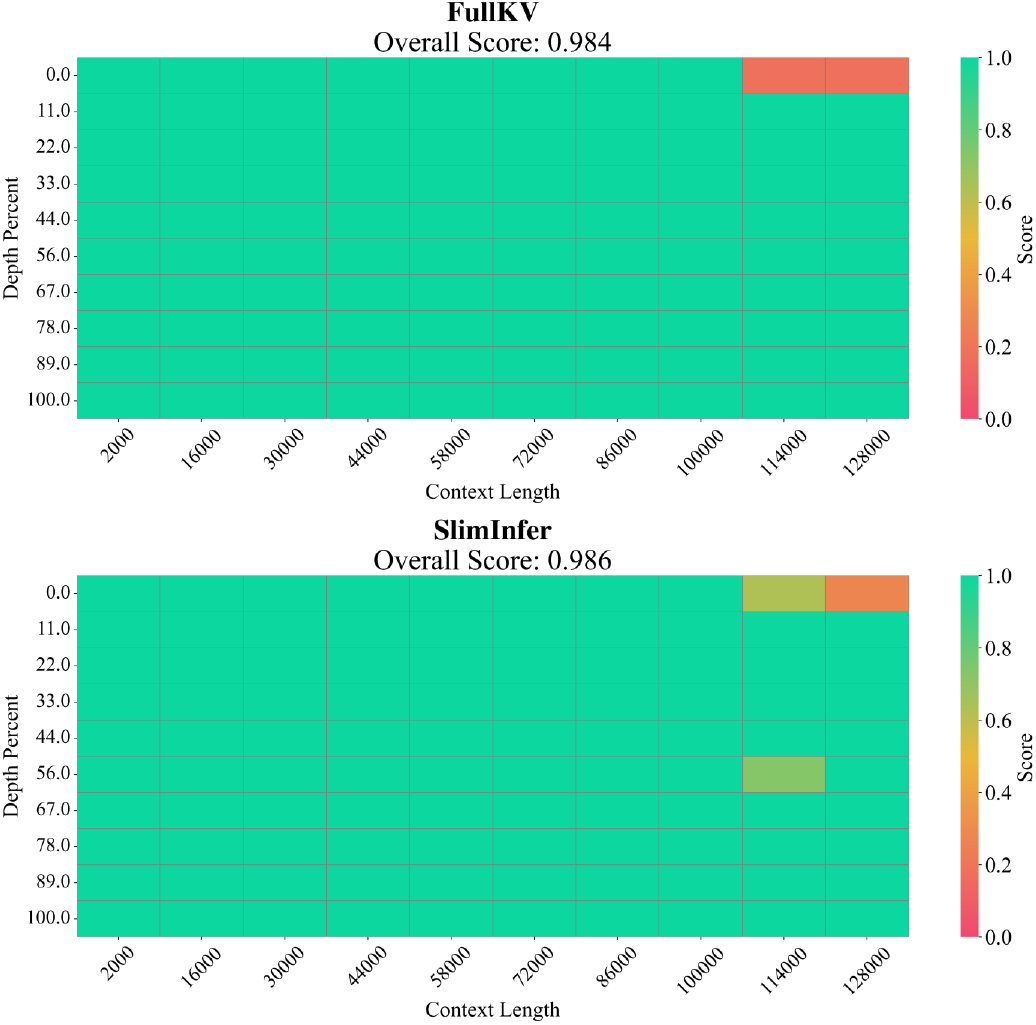} 
\caption{Needle-in-a-Haystack~\cite{kamradt2023niah} test results for Full KV and SlimInfer on LLaMA3.1-8B-Instruct~\cite{grattafiori2024llama3herdmodels}. SlimInfer slightly outperforms the Full KV baseline, demonstrating its ability to enhance retrieval accuracy while accelerating inference.}
\label{fig:niah}
\end{figure}

To assess long-context retrieval, we conduct the Needle-in-a-Haystack~\cite{kamradt2023niah} test using LLaMA3.1-8B-Instruct~\cite{grattafiori2024llama3herdmodels}, with SlimInfer pruning 50\% of tokens at layers 10, 20, and 30. Other configurations remain the same as in the main text. As shown in Figure~\ref{fig:niah}, SlimInfer slightly outperforms the Full KV (Flash Attention2~\cite{dao2023flashattention2fasterattentionbetter}) baseline (0.986 \emph{vs.} 0.984), suggesting that pruning irrelevant tokens reduces noise and helps the model focus on critical information. This demonstrates that SlimInfer not only accelerates inference but can also enhance retrieval accuracy in long contexts.

\section{Additional Efficiency Experiments}
\subsection{Qwen2.5 Latency Profiling}
% In addition to LLaMA3.1-8B-Instruct~\cite{grattafiori2024llama3herdmodels}, we also evaluate the efficiency of SlimInfer on Qwen2.5-7B-Instruct~\cite{qwen2025qwen25technicalreport}, following the same experimental setup. To ensure stable and reliable measurements, all performance results reported in this paper, including those for both models, are averaged over 10 independent runs. The results, presented in Figure~\ref{fig:qwen-efficiency}, demonstrate that SlimInfer consistently outperforms the baselines on the Qwen2.5-7B model as well.

% Specifically, SlimInfer achieves up to \textbf{2.14$\times$} speedup in Time-to-First-Token (TTFT) and \textbf{1.84$\times$} speedup in End-to-End (E2E) latency at a 32k context length. Similar to the results on LLaMA3.1, our method shows a growing acceleration trend as the context length increases, underscoring its effectiveness in long-context scenarios. 

% We observe that the acceleration ratios are slightly lower than those on LLaMA3.1. This is primarily because the pruning configuration for Qwen2.5-7B (retaining 12k, 6k, and 4k tokens) is less aggressive than the one used for LLaMA3.1 (8k, 4k, 2k), resulting in a comparatively lower overall sparsity. Nonetheless, these results validate the strong generalization capability of SlimInfer across different model architectures.

\begin{figure}[h!]
\centering
\includegraphics[width=1.0\columnwidth]{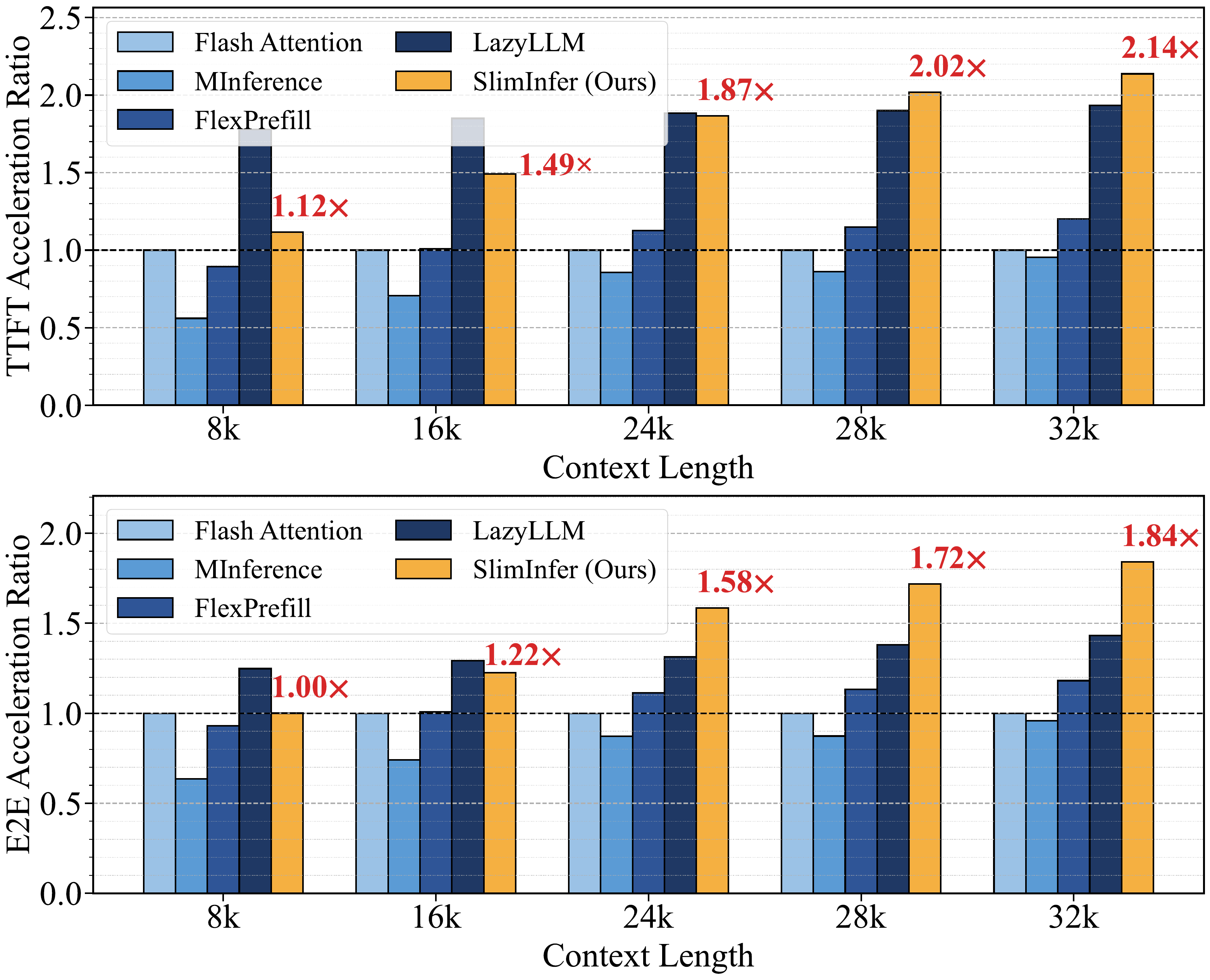} 
\caption{Inference efficiency comparison for Qwen2.5-7B-Instruct~\cite{qwen2025qwen25technicalreport}. (\textbf{Upper}) TTFT and (\textbf{Lower}) E2E latency acceleration ratio \emph{vs.} context length. SlimInfer maintains a significant performance advantage, particularly at longer context lengths.}
\label{fig:qwen-efficiency}
\end{figure}
In addition to LLaMA3.1-8B-Instruct~\cite{grattafiori2024llama3herdmodels}, we also evaluate the performance of SlimInfer on Qwen2.5-7B-Instruct~\cite{qwen2025qwen25technicalreport}, following the same experimental setup. The results, presented in Figure~\ref{fig:qwen-efficiency}, demonstrate that SlimInfer consistently outperforms the baselines on the Qwen2.5-7B-Instruct model as well. Specifically, SlimInfer achieves up to \textbf{2.14$\times$} speedup in Time-to-First-Token (TTFT) and \textbf{1.84$\times$} speedup in End-to-End (E2E) latency at a 32k context length. Similar to the results on LLaMA3.1, our method shows a growing acceleration trend as the context length increases, further underscoring its effectiveness in long-context scenarios. 

% We observe that the acceleration ratios are slightly lower than those on LLaMA3.1. This is primarily because the pruning configuration for Qwen2.5 (retaining 12k, 6k, and 4k tokens) is less aggressive than the one used for LLaMA3.1 (8k, 4k, 2k), resulting in a comparatively lower overall sparsity. Nonetheless, these results validate the strong generalization capability of SlimInfer across different model architectures.

\subsection{Details for Accuracy \emph{vs.} Efficiency Analysis}

\begin{table*}[ht!]
\centering
\begin{tabular}{lcccccc}
\toprule
\multirow{2}{*}{Context} & \multicolumn{2}{c}{Full KV (Baseline)} & \multicolumn{2}{c}{SlimInfer (Ours)} & \multicolumn{2}{c}{Speedup} \\
\cmidrule(r){2-3} \cmidrule(r){4-5} \cmidrule(l){6-7}
& TTFT   & E2E & TTFT   & E2E & TTFT & E2E \\
\midrule
4k  & 12.49 & 18.22 & 12.76 & 19.56 & 0.98$\times$ & 0.93$\times$ \\
8k  & 26.16 & 34.51 & 22.78 & 33.34 & 1.15$\times$ & 1.04$\times$ \\
16k & 57.05 & 72.71 & 35.31 & 59.38 & 1.62$\times$ & 1.22$\times$ \\
24k & 93.31 & 115.60 & 51.01 & 85.62 & 1.83$\times$ & 1.35$\times$ \\
28k & 113.35 & 138.62 & 59.29 & 86.58 & 1.91$\times$ & 1.60$\times$ \\
32k & 134.73 & 163.10 & 69.57 & 96.06 & \textbf{1.94$\times$} & \textbf{1.69$\times$} \\
\bottomrule
\end{tabular}
\caption{Performance comparison on NVIDIA Jetson AGX Orin (32GB) using LLaMA3.1-8B-Instruct~\cite{grattafiori2024llama3herdmodels}. Latency is measured in seconds. The end-to-end (E2E) latency corresponds to the time taken to generate 16 tokens.}
\label{tab:jetson_performance}
\end{table*}

This section details the specific configurations for the accuracy \emph{vs.} efficiency trade-off results (\emph{i.e.}, the visualization of Pareto frontier in the main text), as presented comprehensively in Table~\ref{tab:longbench_appendix_llama}. For FlexPrefill, we varied its sparsity threshold $\gamma$ across \{0.99, 0.98, 0.95, 0.90\}. For LazyLLM, we adjusted the token retention ratio through the set \{50\%, 45\%, 40\%, 35\%\}. For our method, SlimInfer, we tested four pruning schedules at layers 10, 20, and 30, denoted by the number of retained tokens (in units of k, where $1\text{k} = 1024$): (16, 8, 4), (12, 6, 4), (10, 5, 3), and (8, 4, 2). The (8, 4, 2) setting represents our most aggressive configuration and corresponds to the main results in the paper. Other baselines like Full KV and MInference were run with their default settings. The results in Table~\ref{tab:longbench_appendix_llama} illustrate the direct relationship between computational reduction and model performance. As expected, more aggressive configurations in FlexPrefill, LazyLLM, and SlimInfer lead to lower E2E latency. Notably, SlimInfer's configurations consistently offer a more favorable balance, achieving substantial latency reductions while incurring minimal accuracy degradation, effectively pushing the Pareto frontier.

\subsection{Performance on Edge Devices}

To further validate the practicality and broad applicability of SlimInfer, we conducted additional performance evaluations on an edge computing platform, the NVIDIA Jetson AGX Orin (32GB). Due to limitations of the Triton~\cite{10.1145/3315508.3329973}, which is not supported on the Jetson platform, we were unable to run other baselines such as MInference and FlexPrefill. Therefore, we directly compare SlimInfer against a highly-optimized Full KV baseline. For this experiment, we used the LLaMA3.1-8B-Instruct model with the (8, 4, 2) pruning configuration for SlimInfer. The results are summarized in Table~\ref{tab:jetson_performance}. SlimInfer demonstrates substantial performance gains over the Full KV baseline across all context lengths. Notably, at a 32k context length, SlimInfer achieves a \textbf{1.94$\times$} speedup in Time-to-First-Token (TTFT) and a \textbf{1.69$\times$} speedup in End-to-End (E2E) latency for generating 16 tokens. These results indicate that SlimInfer is not only effective on high-end server GPUs but also provides significant acceleration on edge devices, highlighting its excellent generalization and practical value for real-world deployment.

\section{Additional Ablation Study}
We use LLaMA3.1‑8B‑Instruct~\cite{grattafiori2024llama3herdmodels} here. The default settings are given in the main text.
\subsection{Ablation Study on Swap Threshold}
\begin{table}[ht!]
\centering
\setlength{\tabcolsep}{4pt}
\begin{tabular}{ccc}
\toprule
Threshold ($\gamma$) & Avg. Score (\%) & E2E (s) \\
\midrule
0.99        & 47.67 & 3.057 \\
0.95        & 47.67 & 3.040 \\
0.90        & 47.65 & 2.959 \\
\bottomrule
\end{tabular}
\caption{Ablation on the swap threshold ($\gamma$) for LLaMA3.1-8B-Instruct~\cite{grattafiori2024llama3herdmodels}. We report the average score on LongBench~\cite{bai2024longbenchbilingualmultitaskbenchmark} and E2E latency (16 tokens generated, 32k input).}
\label{tab:ablation_threshold}
\end{table}

\label{sec:appendix_ablation_threshold}

We study the impact of the swapping threshold, $\gamma$, which controls the frequency of asynchronous data transfers. A lower $\gamma$ reduces I/O overhead by allowing more tolerance for changes in the active token set. As shown in Table~\ref{tab:ablation_threshold}, decreasing the threshold from 0.99 to our default value of 0.90 reduces the E2E latency from 3.057s to 2.959s. This performance gain comes at the cost of a negligible 0.02\% drop in LongBench~\cite{bai2024longbenchbilingualmultitaskbenchmark} average score. This favorable trade-off justifies our choice of $\gamma=0.90$ to optimize for inference speed with minimal impact on accuracy.

\subsection{Ablation Study on Block Size}
\label{sec:appendix_ablation_block_size}

\begin{figure}[!ht]
\centering
\includegraphics[width=1.0\columnwidth]{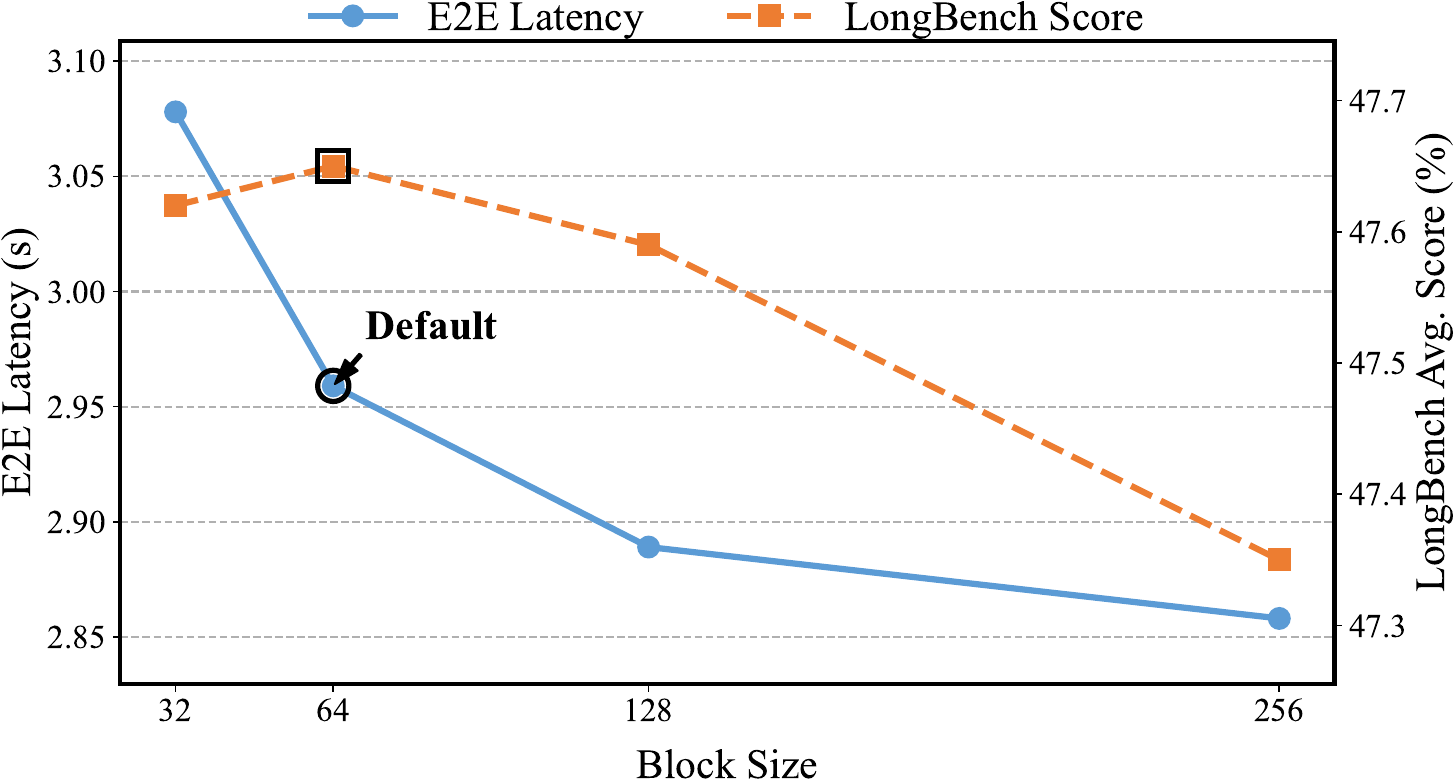} 
\caption{Ablation study on block size, showing the LongBench~\cite{bai2024longbenchbilingualmultitaskbenchmark} average score and end-to-end (E2E) latency under different block size settings. }
\label{fig:ablation_block_size}
\end{figure}

We study the effect of block size on both performance and accuracy. As shown in Figure~\ref{fig:ablation_block_size}, increasing the block size consistently reduces E2E latency, measured under a 32k token input and 16 token output setting. This reduction stems from improved computational and memory efficiency on the GPU, as larger blocks reduce the number of memory operations and pruning decisions required. However, the impact on accuracy is non-monotonic. When the block size is too small (\emph{e.g.}, 32), each block contains limited context, making it difficult to capture coherent semantic patterns, which undermines the accuracy of importance evaluation. As the block size increases, accuracy improves and peaks at block size 64, where local semantic structure is preserved without introducing too much noise. Beyond this point, further increases in block size lead to performance degradation, as overly large blocks tend to incorporate irrelevant tokens, diluting critical information and reducing pruning precision.

\subsection{Ablation Study on Local Query Window Size}
\begin{figure}[!ht]
\centering
\includegraphics[width=1.0\columnwidth]{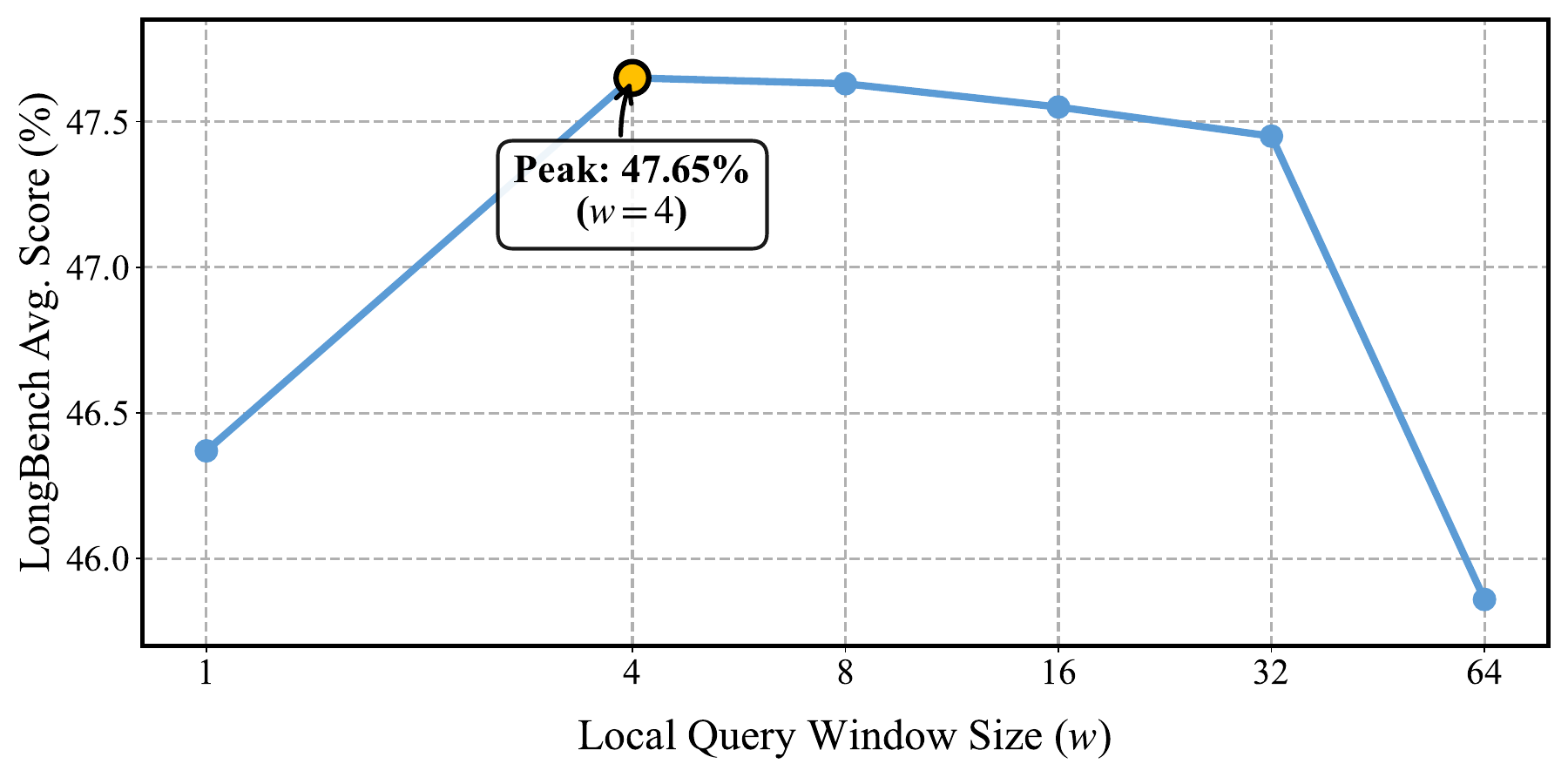} 
\caption{Ablation study on the local query window size ($w$), showing the average score on LongBench~\cite{bai2024longbenchbilingualmultitaskbenchmark} under different window size settings.}
\label{fig:ablation_window_size}
\end{figure}

We perform an ablation study on the local query window size ($w$), which determines the number of recent tokens used to construct the local Query vector for importance scoring. This parameter plays a crucial role in accurately identifying relevant prompt blocks based on the current semantic context. As shown in Figure~\ref{fig:ablation_window_size}, the model's performance is highly sensitive to the choice of window size. The average score on LongBench~\cite{bai2024longbenchbilingualmultitaskbenchmark} reaches its peak at $\mathbf{w=4}$. A smaller window size (\emph{e.g.}, $w=1$) results in a notable performance drop, likely due to insufficient contextual information for reliable importance estimation. In contrast, excessively large windows (\emph{e.g.}, $w \geq 8$) also lead to a gradual decrease in precision, suggesting that they may incorporate outdated or irrelevant tokens, thus diluting the effectiveness of the importance score. Overall, a window size of 4 offers an optimal balance by capturing a sufficiently stable and relevant context while minimizing the influence of semantic noise.

% \bigskip
% \noindent Thank you for reading these instructions carefully. We look forward to receiving your electronic files!

\ifappendixonly
\bibliography{aaai2026}
\else
\fi

% \newpage
% \input{../../ReproducibilityChecklist/LaTeX/ReproducibilityChecklist.tex}

\end{document}